\documentclass[conference,a4paper]{IEEEtran}

\usepackage{cite}
\usepackage{amsmath,amssymb,amsfonts}
\usepackage{algorithmic}
\usepackage{graphicx}
\usepackage{caption}
\usepackage{subcaption}
\usepackage[linesnumbered,ruled]{algorithm2e}
\usepackage{textcomp}
\usepackage{xcolor}
\usepackage{romannum}
\usepackage{graphicx}
\usepackage{flushend}
\def\BibTeX{{\rm B\kern-.05em{\sc i\kern-.025em b}\kern-.08em
    T\kern-.1667em\lower.7ex\hbox{E}\kern-.125emX}}
\begin{document}

\title{Finding neural signatures for obesity through feature selection on source-localized EEG  }

\author{\IEEEauthorblockN{Yuan Yue}
\IEEEauthorblockA{\textit{Department of Information Science} \\
\textit{University of Otago}\\
Dunedin, New Zealand \\
{\small\tt yueyu445@student.otago.ac.nz}}
\and
\IEEEauthorblockN{Dirk De Ridder}
\IEEEauthorblockA{\textit{Department of Surgical Science} \\
\textit{University of Otago}\\
Dunedin, New Zealand \\
{\small\tt dirk.deridder@otago.ac.nz}}
\and
\IEEEauthorblockN{Patrick Manning}
\IEEEauthorblockA{\textit{Department of Medicine} \\
\textit{University of Otago}\\
Dunedin, New Zealand \\
{\small\tt patrick.manning@otago.ac.nz}}
\and
\IEEEauthorblockN{Matt Hall}
\IEEEauthorblockA{\textit{Department of Surgical Science} \\
\textit{University of Otago}\\
Dunedin, New Zealand \\
{\small\tt matt.hall@postgrad.otago.ac.nz}}
\and
\IEEEauthorblockN{Samantha Ross}
\IEEEauthorblockA{\textit{Department of Medicine} \\
\textit{University of Otago}\\
Dunedin, New Zealand \\
{\small\tt samantha.ross@otago.ac.nz}}
\and
\IEEEauthorblockN{Jeremiah D. Deng}
\IEEEauthorblockA{\textit{Department of Information Science} \\
\textit{University of Otago}\\
Dunedin, New Zealand \\
{\small\tt jeremiah.deng@otago.ac.nz}}
}
\maketitle

\begin{abstract}
Obesity is a serious issue in the modern society and is often associated to significantly reduced quality of life. Current research conducted to explore obesity-related neurological evidences using electroencephalography (EEG) data are limited to traditional approaches. In this study, we developed a novel machine learning model to identify brain networks of obese females using alpha band functional connectivity features derived from EEG data. An overall classification accuracy of 0.937 is achieved. Our finding suggests that the obese brain is characterized by a dysfunctional network in which the areas that responsible for processing self-referential information and environmental context information are impaired.
\end{abstract}

\section{Introduction}


Obesity is a a global health problem today which is associated with various health risks ~\cite{volkow2011}. In recent years, there has been a growing interest in understanding the neural underpinnings of obesity and its potential impact on brain function. Particularly, abnormal functional connectivity (FC) patterns (i.e., interactions between brain regions which may disrupt normal brain network communication) in obese individuals are extensively investigated ~\cite{kullmann2016,gupta2015,syan2021}. 

Several studies have consistently reported alterations in resting-state FC within the brain networks associated with reward processing, emotional regulation, and cognitive control in obese individuals ~\cite{syan2021,mcintyrewood2019,brucekeller2010}. 
Moreover, studies have identified abnormal connectivity patterns between the hypothalamus and brain regions involved in reward, limbic processing, and cognitive functions, such as the limbic system, prefrontal cortex, and insula ~\cite{liang2016,carnell2014}. These alterations may play a crucial role in the dysregulation of appetite and energy homeostasis, contributing to the development and progression of obesity.

However, to our best knowledge, existing studies investigate obesity-related neurological evidences use traditional statistical approaches that come with limitations ~\cite{li2019}. First, traditional statistical approaches rely on predefined brain regions or networks of interest, potentially missing important connections. They may also struggle to handle the complexity and non-linear relationships within neurological data. To overcome these limitations and to explore complex interactions between brain regions, the integration of machine learning approaches is needed.

Our study is the first one that aims to use an EEG-based data-driven machine learning approach to identify the obese brain network with minimal yet necessary brain connections (i.e., FC). Then by exercising targeted intervention on these connections in the clinical context, obesity-related behaviors that are caused by neural factors can be improved.
Specifically, we developed a novel feature selection scheme named MICMAC, which can effectively extract a few features that significantly contribute to the classification performance from high-dimensional input datasets.


\section{Data Acquisition}
\label{sec:data}
Thirty obese females and thirty lean (i.e., healthy) females between 25 to 65 years old participated in our study.  Subjects who have a Body Mass Index (BMI) higher than 30 are defined as obese individuals, and those with BMI lower than 25 are defined as lean individuals.
Participants who have obesity associated co-morbidities are excluded from this study.
Details regarding participants recruitment can be found in ~\cite{ross2018quantitative}.

\begin{figure}[!t]
    \centering
    \includegraphics[width=\columnwidth]{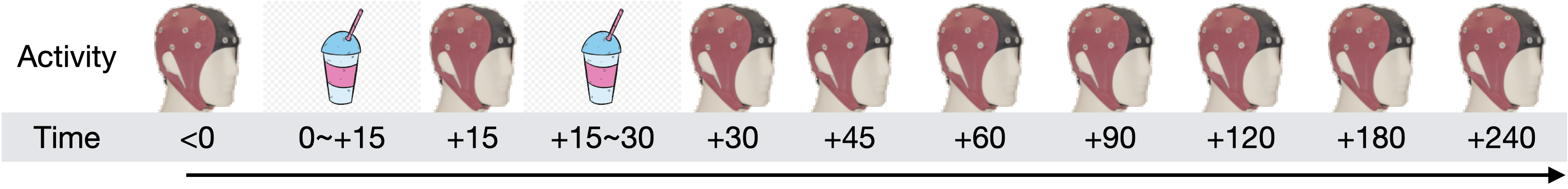}
    \caption{The  schedule with 2 meals and 9 time-points (in minutes) for EEG capture.}
    \label{fig:timepts}
\end{figure}

EEG data used in our study were recorded through the international 10–20 system. During the recording sessions, the acquisition environment 
was controlled to be in the same condition for all the subjects. 
EEG data are recorded at 9 time-points during eye-closed resting state. The first EEG dataset was captured when all subjects were on empty stomachs. A liquid meal followed. The second EEG was captured 15 minutes after the meal. A second liquid meal followed. Then 7 more EEG captures occurred at various time points after the second meal. The schedule is illustrated in Fig.~\ref{fig:timepts}.  

\section{Method}
\label{sec:Method}

\subsection{Data Pre-processing }\label{AA}
The first five seconds of each EEG recording were discarded as they usually contain a high level of noise. The EEG recordings were then resampled to 128Hz, and band-pass filtered to 0.1Hz - 45Hz. We then cut each EEG time series into consecutive 20-second segments.

\subsection{Source Localization and Feature Extraction}
Source localization was performed using the Low-resolution Electromagnetic Tomographic Analysis (LORETA) ~\cite{pascual-marqui_assessing_2011}. The Regions of interest (ROI) selected for the process were based on the Brodmann brain atlas, which has been widely used to represent the localization of brain function.
We further divided the Brodmann areas no.53 and no.54 into 3 sub regions to improve the resolution of the brain map. Totally, 88 ROIs are localized for further feature computation.

It is suggested that the lagged components of FC (i.e. lagged coherence) are minimally affected by the low scalp spatial resolution, thus containing high quality physiological information ~\cite{pascual-marqui_assessing_2011}. Based on initial evaluation across different frequency bands, the lagged coherence features in the alpha band (8-13Hz) were employed in our study for obesity classification.
Between the 88 ROIs, $88\times 87/2=3828$ distinct lagged coherence features are computed. 

\subsection{Subject-based Cross Validation }
The lagged coherence feature set with a size of 540 (9 time points $\times$ 60 subjects = 540) $\times$ 3828 is split for  training and testing using subject-based 10-fold cross validation. In each of the 10 folds, we select 54 subjects’ data for training and validating, and 6 subjects’ data for testing. The training and validating data are then further divided using subject-based 9-fold cross validation (each fold consists of 48 training subjects and 6 validating subjects). 

\subsection{Feature Selection }
The feature selection method proposed in this study involves several key steps explained as follows.

\subsubsection{Feature preselection}
\label{step1}
We first reduced the size of the feature set using a random forest classifier with ``max depth''  set to 10. The top 100 features with the highest importance scores were then selected for further process.

\subsubsection{Feature selection}
In this step, we propose a novel, Maximal Incremental Contribution with Minimal Average Correlation (MICMAC) method for feature selection. Denote the reduced feature set by $F_0$. We start with an initial selection $F_1=\{f_1\}$, where $f_1$ is the feature in $F_0$ that has the highest importance. 
Our goal is to select features using a classifier (i.e., wrapper), assessing the merit scores of all the other features not yet selected, by looking at their potential gain in classification performance versus their correlation with the existing feature selection and selecting the best features incrementally. This gives us the best feature addition that maximizes performance gain while minimizing redundancy with selected features. 
Specifically, suppose at the $m$-th step we have selected feature set $F_m$. We define the merit score for a feature candidate $f_i \in F_0\backslash F_m$ as:
\begin{equation}
    \mu(f_i, F_m)=\frac{\phi(X(F_{m}\cup f_i))-\phi(X(F_{m}))}{\sum_{f_k\in F_m} \mathrm{cosine}(\mathbf{x}(f_i),\mathbf{x}(f_k))},
\label{eq:featscore}
\end{equation}
where $X(F)$ indicates the dataset using feature set $F$, $\mathbf{x}(.)$ stands for the data vector of a particular feature, $\mathrm{cosine}(.)$ gives the cosine similarity between two vectors as a redundancy measure, and $\phi(.)$ gives the classification accuracy obtained from one fold of validation. 
We then pick the best feature with the top score:
\begin{equation}
f_s = \arg\max_{f_i\in F_0\backslash F_{m}}\mu(f_i, F_{m}),
\label{eq:feateval_ratio}
\end{equation} 
and update the feature selection by
\begin{equation}
    F_{m+1}=F_{m}\cup f_s,
\end{equation}
if the best merit score is higher than a threshold $T$: $\mu(f_s, F_m)>T$; otherwise we terminate the feature selection procedure. The MICMAC algorithm is given in Algorithm~\ref{algo:mpgmc}. 

The algorithm is iterated on each of the training and validation folds.
Next,  we re-rank all the selected features in a descending order according to the number of times they are selected across all the folds. 

Although our algorithmic framework has high flexibility and interpretability in terms of algorithmic choices, we chose K-nearest neighbors (KNN) with K=3 and support vector machine (SVM) with a Radial Basis Functional kernel as the wrapper of the MICMAC algorithm.

\subsubsection{Classification }
We also used the KNN and SVM classifiers for final classification. 
Since each subject has 9 samples obtained from different time points, we then determined the final label of each subject using a simple majority vote on the classification results on the samples. 

\begin{algorithm}[!t]
\caption{MICMAC Feature Selection}
\label{algo:mpgmc}
\KwData{Initial feature set $F_0$, threshold $T$}
\KwResult{Selected feature set $F$}
Assign the best feature to selection: $F \leftarrow f_1$\;
\For{$f_i$ in $F_0 \backslash F$}
{
   Calculate $\mu(f_k, F)$, $\forall f_k\in F_0 \backslash F$\ using Eq.~(\ref{eq:featscore})\;
   Pick the top-score feature $f_s$ according to Eq.~(\ref{eq:feateval_ratio})\;
   \eIf {$\mu(f_s, F)>T$} {
      $F\leftarrow F\cup f_s$\;
    }
    {
    break\;
    }
 }
Return $F$\;
\end{algorithm}

\begin{figure*}[!t]
    \centering
    \includegraphics[width=0.78\textwidth]{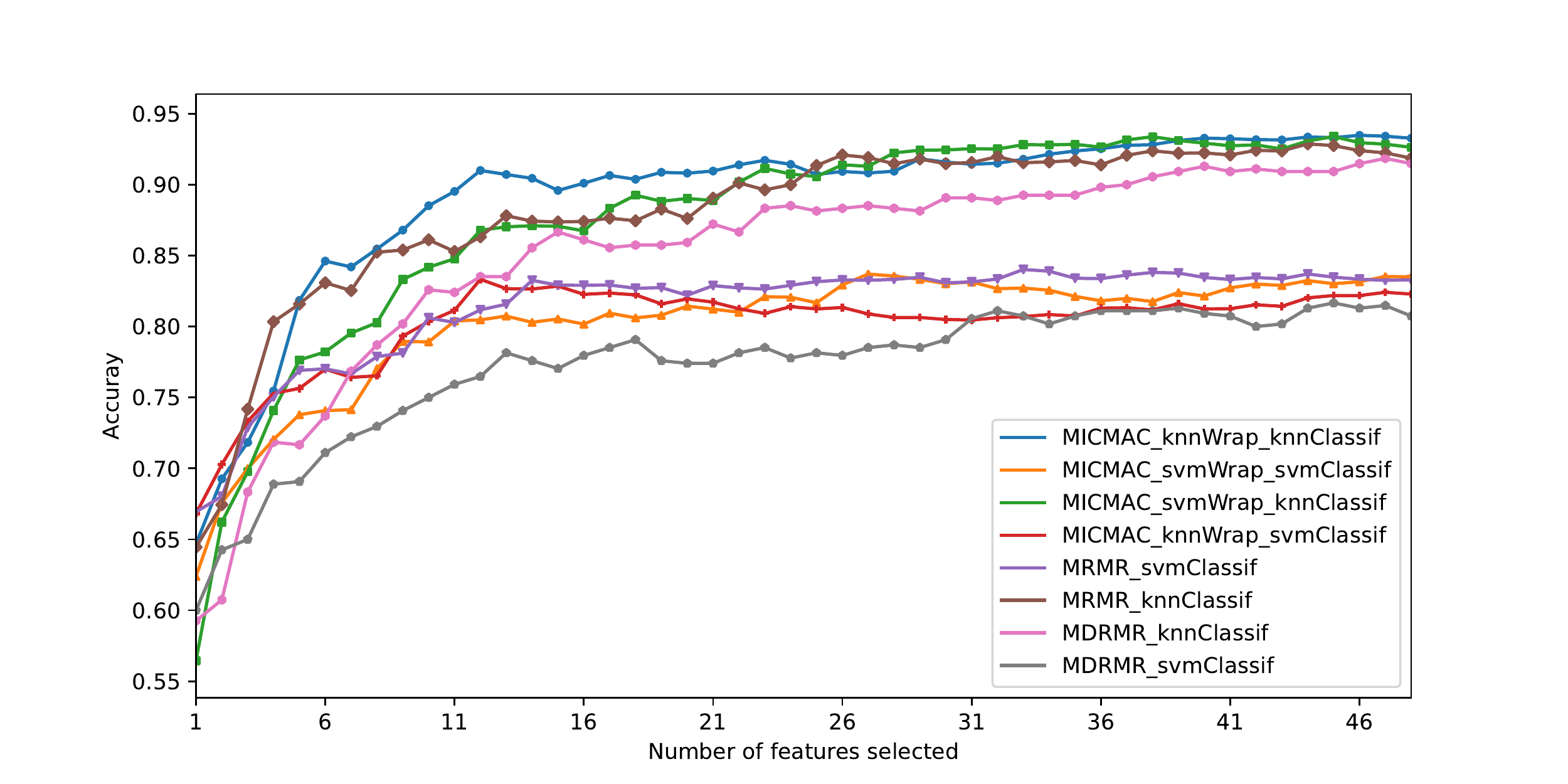}
    \caption{Averaged accuracy with different number of features selected.}
    \label{fig:numb_feat}
\end{figure*}
\subsubsection{Running the experiment multiple times  }
The above process including feature selection through multi-fold validation and classification is called an ``experiment''. To improve the reliability of our results, we then conducted 10 experiments, and we randomly shuffled the dataset each time.

\section{Results and Discussion }
\label{sec:res}

 Fig.~\ref{fig:numb_feat} shows the MICMAC performances with all wrapper and classifier combinations when different numbers of features are selected. We also compared MICMAC with a widely used feature selection method named Min-Redundancy Max-Relevance (mRMR) and a state-of-the-art method developing based on mRMR named Max Dynamic Relevancy and Min Redundancy (MDRMR) ~\cite{peng2005feature,Yin2023}. 
 

The averaged test score (accuracy) of each model with the optimal feature selection is shown in Table ~\ref{table:score} (``Best Selection''). The best classification performance is obtained when using KNN as the classifier while using either a KNN-based MICMAC or a SVM-based MICMAC to select features (accuracy = 0.937). However, we argue that the KNN-based MICMAC outperforms the SVM-based MICMAC in spite of the same test scores are obtained as it is shown in Fig.~\ref{fig:numb_feat} that a relatively good model performance (accuracy=0.902) can be achieved by using only 12 features when a KNN-based MICMAC is applied. To further demonstrate this, we calculated the test scores with all different wrapper and classifier combinations when 12 features are selected, these scores are demonstrated in Table ~\ref{table:score}. It can be seen that with such a limited number of features selected, a KNN-based MICMAC with a KNN classifier outperforms all other models. 

\begin{table}[htb]
    \centering
    \begin{tabular}{|l|c|c|} \hline
         Scheme & Best Selection & Top-12 Selection \\ \hline\hline
         MICMAC-knnW-knnC & \textbf{0.937$\pm$0.199} (50) & \textbf{0.902$\pm$0.039} \\
         MICMAC-knnW-svmC & 0.813$\pm$0.038 (12) & 0.813$\pm$0.038 \\
         MICMAC-svmW-knnC & \textbf{0.937$\pm$0.199} (38) & 0.859$\pm$0.051 \\          MICMAC-svmW-svmC & 0.822$\pm$0.045 (38) & 0.813$\pm$0.033 \\ \hline
         mRMR-knnC & 0.917$\pm$0.040 (26) & 0.867$\pm$0.050 \\
         mRMR-svmC & 0.819$\pm$0.036 (14) & 0.809$\pm$0.052 \\ \hline
         MDRMR-knnC & 0.909$\pm$0.039 (47) & 0.826$\pm$0.029 \\
         MDRMR-svmC & 0.802$\pm$0.043 (32) & 0.754$\pm$0.056 \\ \hline
    \end{tabular}
    \caption{Testing performance compared across different feature selection algorithms using varying wrapper/classifier options: ``W'' after an algorithm indicates its use as a wrapper, ``C'' as a classifier. Numbers in brackets give the numbers of features given by the feature selection algorithms for the best outcome. Overall best outcomes are in bold. }
\label{table:score}
\end{table}

We further conducted the Tukey's Honestly Significant Difference test to compare whether the classification performance between using the 12 features selected by KNN-based MICMAC and the 38 features selected by SVM-based MICMAC, and between using the 12 features selected by KNN-based MICMAC and the 26 features selected by mRMR, are significantly different (all feature selection methods tested here are on the KNN classifier). The $p$-values of 0.678 and 0.976 for the two pairs of comparison, respectively, suggest that KNN-based MICMAC selected significantly less number of features while achieving a classification performance that is not significantly different from the other two top-performing models.



\begin{figure}
    \centering
    \includegraphics[width=0.6\columnwidth]{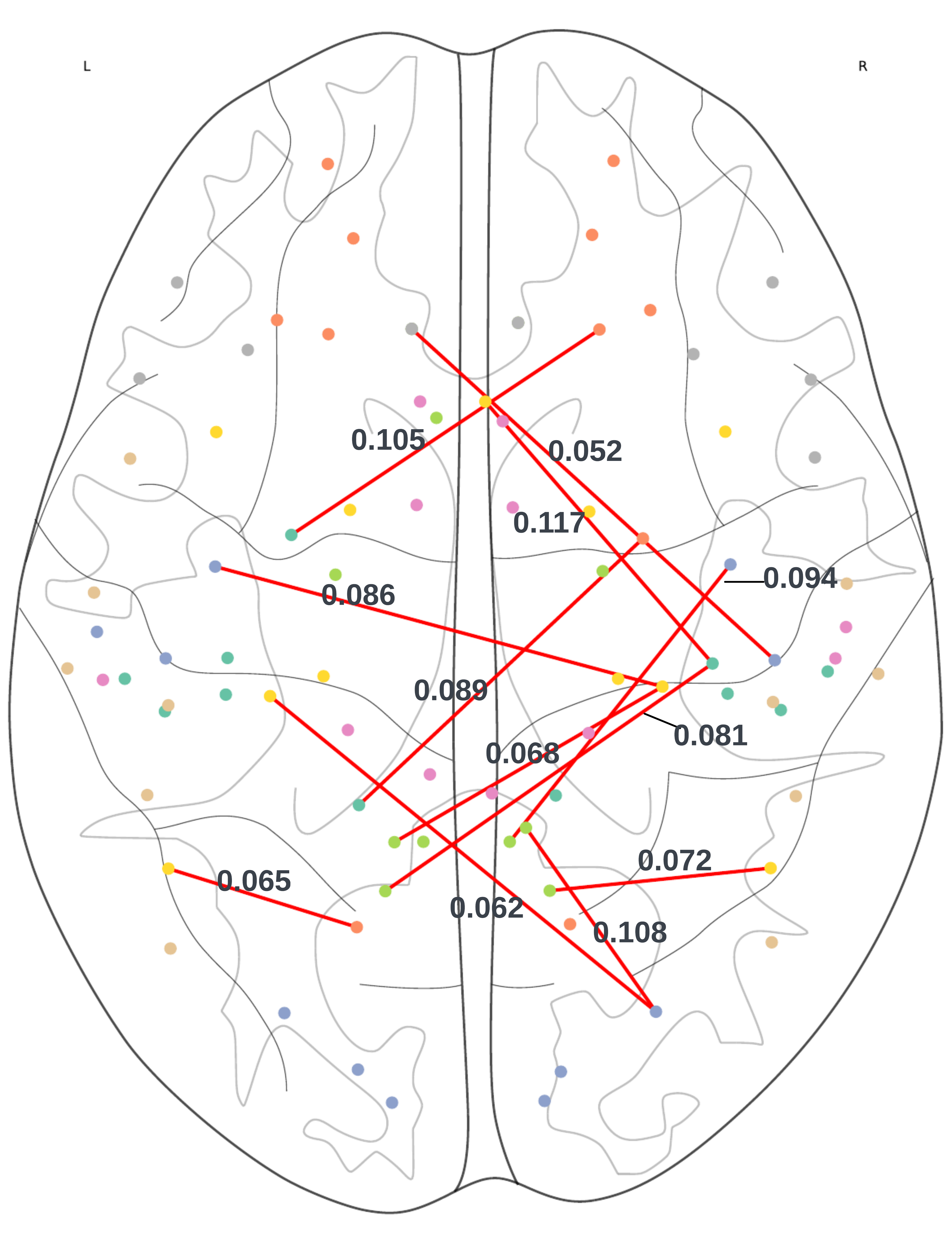}
    \caption{The 12 FC features selected by KNN-based MICMAC.}
    \label{fig:12f}
\end{figure}

The 12 selected features using KNN-based MICMAC are shown in Fig.~\ref{fig:12f}. Each line represents a lagged coherence connection between two brain regions. We also computed the feature importance scores using a random forest classifier, represented by the values above each line. From the neurological perspective, alpha band communication between the somatosensory cortex which is responsible for encoding the exteroceptive self (i.e., how we implicitly perceive our body in relation to environmental context) and insula which functions to encode the interceptive self (i.e., how we unconsciously processing inner sensations) as well as  to track energy need, is observed in obese brain~\cite{menon_saliency_2010,brooks_increased_2013}. This suggests that obese women are potentially less capable of predicting their energy intakes. 
In addition, abnormalities are observed in the posterior cingulate cortex area, the temporoparietal and inferior parietal area, and the parahippocampal area in obese women. The connection between the temporo-occipital junction and the parahippocampal area which can be observed in healthy brain is also missing in obese brains. Since these areas works together for indivuals to process environmental cues and adapt their behaviors accordingly, we can then infer that contextual information may influence obese women less than lean women, potentially explaining the impaired ability of predicting energy requirement of obese women ~\cite{pearson_posterior_2011,by_accurate_2020}.

Our findings are in line with the previous research discussed in Section I. Moreover, our study further reduces the complexity of obese brain network by identifying the most few important distinctive brain connections as describe above. The analyzation cost of future obesity-related studies can be reduced by focusing on the few important brain connections descried here rather than the entire massive brain network.

\section{Conclusion}
\label{sec:conc}
In this study, we have investigated obesity-related neurological characteristics in resting-state EEG signals using a machine learning approach. By using a new feature selection scheme MICMAC we effectively reduced the large dimensionality of the input dataset. Distinctive brain connection patterns were found between obese females and lean females.  The distinction primarily associates to the impairment of brain areas which function to encode self-referential information and to evaluate energy requirement. Neuromodulation treatments can be developed upon these dintinctive brain connections to improve obesity-related behaviours.                                     

For future work, we intend to collect more EEG data from subjects of both sexes and apply our method to identify reliable diabetes neural signatures, which may pave the way for effective EEG-based diagnosis as well as interventions. 

\bibliographystyle{IEEEtran}
\bibliography{Bibliography}

\end{document}